\def\year{2022}\relax
\documentclass[letterpaper]{article} % DO NOT CHANGE THIS
\usepackage{aaai22}  % DO NOT CHANGE THIS
\usepackage{times}  % DO NOT CHANGE THIS
\usepackage{helvet}  % DO NOT CHANGE THIS
\usepackage{courier}  % DO NOT CHANGE THIS
\usepackage[hyphens]{url}  % DO NOT CHANGE THIS
\usepackage{graphicx} % DO NOT CHANGE THIS
\usepackage{natbib}  % DO NOT CHANGE THIS AND DO NOT ADD ANY OPTIONS TO IT
\usepackage{caption} % DO NOT CHANGE THIS AND DO NOT ADD ANY OPTIONS TO IT
\DeclareCaptionStyle{ruled}{labelfont=normalfont,labelsep=colon,strut=off} % DO NOT CHANGE THIS
\usepackage{algorithm}
\usepackage{algorithmic}

\usepackage{microtype}
\usepackage{subfigure}
\usepackage{booktabs} % for professional tables
\usepackage{amsmath}
\usepackage{multirow}

\usepackage{listings}
\usepackage{color}
\usepackage{xcolor}
\definecolor{dkgreen}{rgb}{0,0.6,0}
\definecolor{gray}{rgb}{0.5,0.5,0.5}
\definecolor{mauve}{rgb}{0.58,0,0.82}
\title{Graph Conditioned Sparse-Attention for Improved Source Code Understanding}

\author {
    Junyan Cheng,
    Iordanis Fostiropoulos,
    Barry Boehm
}
\affiliations {
    University of 
Southern California\\Los Angeles, California 90007\\
    junyanch@usc.edu, fostirop@usc.edu, boehm@usc.edu
}

\usepackage{bibentry}
\begin{document}

\maketitle

\begin{abstract}
Transformer \cite{Transformer} architectures have been successfully used in learning source code representations. The fusion between a graph representation like Abstract Syntax Tree (AST) and a source code sequence makes the use of current approaches computationally intractable for large input sequence lengths. Source code can have long-range dependencies that require larger sequence lengths to model effectively. Current approaches have a quadratic growth in computational and memory costs with respect to the sequence length. Using such models in practical scenarios is difficult.
In this work, we propose the conditioning of a source code snippet with its graph modality by using the graph adjacency matrix as an attention mask for a sparse self-attention mechanism and the use of a graph diffusion mechanism to model longer-range token dependencies. 
Our model reaches state-of-the-art results in BLEU, METEOR, and ROUGE-L metrics for the code summarization task and near state-of-the-art accuracy in the variable misuse task. The memory use and inference time of our model have linear growth with respect to the input sequence length as compared to the quadratic growth of previous works. 

\end{abstract}

\begin{figure*}[h]
\begin{center}
\includegraphics[width=1\textwidth]{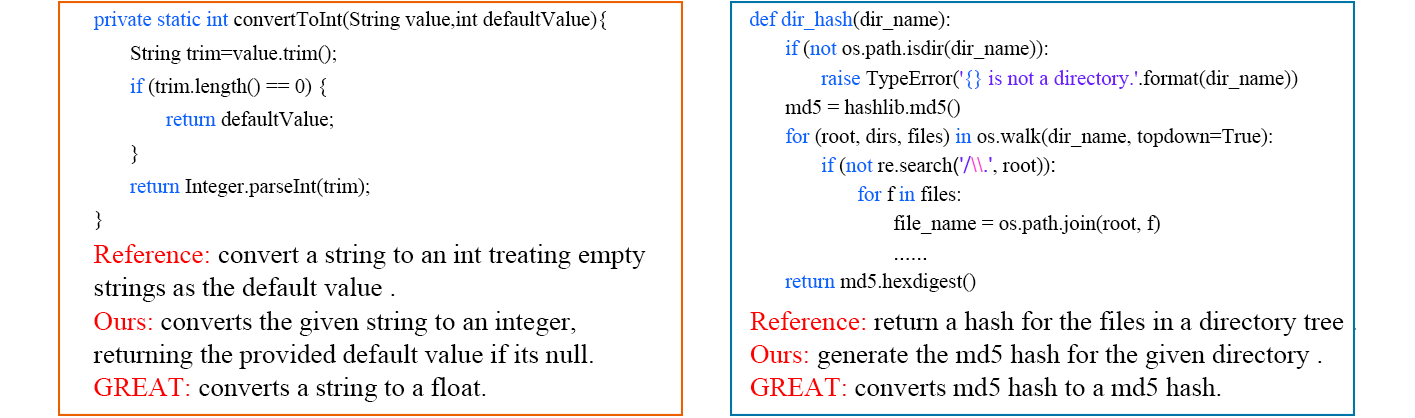}
\end{center}
\caption{Examples of our model evaluated on the code summarization task on Java (red block) and Python (blue block) test set. GREAT refers to the work by \cite{GREAT}}
\label{fig:cssamples}
\end{figure*}

\noindent Traditional program static analysis techniques generate rich graph representations for code. Such representations can contain valuable information that is being successfully used for automatic code analysis techniques.
For example, \citet{flowdroid} applied inter-procedural Control Flow Graphs (CFG) in the taint analysis for Android apps. \citet{infer} used call graphs in the analysis of memory safety for C programs. 
How to effectively incorporate those representations has been one of the core challenges for learning distributed representations of code \cite{learn_code_graph}. 
Transformers have succeeded in multiple natural language understanding tasks, and have also been evaluated on the understanding of programming languages \cite{Ahmad}.
A lot of efforts have been made on incorporating code graph representations into a Transformer's architecture. \citet{Ahmad} use Structure-Based Traversal (SBT) to serialize Abstract Syntax Trees (AST) as input to a Transformer. However, they observed an accuracy decrease for the code summarization task. 
GREAT \cite{GREAT} uses a relation-aware self-attention mechanism to model edge information as the relationship between tokens. 
Combined with a fused graph representation, \citet{GREAT} achieved state of art in the variable misuse task. \citet{CodeTransformer} extract relative positional information from AST structure to improve accuracy in code summarization.
 
There are two shortcomings of Transformers architecture, the first is that it makes no assumptions on the inductive biases of the data, thus it requires a large amount of training data to outperform a model with good inductive biases as shown by recent works \cite{VisionTransformer}. 
A way of introducing inductive bias is through a graph structure, a graph structure brings relational inductive biases among input tokens \cite{GN}. Inductive bias may improve the \textit{sample efficiency} \cite{GN} and thus improve performance with a smaller size dataset. 

The second shortcoming is that the computational complexity of the attention matrix is quadratic with respect to the sequence length.
The above methods have achieved good accuracy by introducing inductive bias from a graph into a Transformer architecture. However, they are computationally inefficient as the sequence length increases. They are evaluated on snippets of limited length (e.g. 150, 400 tokens for Java and Python dataset, and 512 for VarMisuse dataset). In the context of a source code file or software system, the sequence length is minuscule in capturing longer-range dependencies that are naturally present. 
We found that, for popular repositories such as Tensorflow, the average number of lines per file is 208.4, additional metrics on popular GitHub projects could be found in Appendix \ref{apdx:gitproj}.
The quadratic growth with respect to the sequence length in a Transformer architecture \cite{SparseTransformer}, in combination with the need for longer sequence lengths for improved model accuracy, makes the current works difficult to be applied in a practical setting. 

In recent years, there have been many methods proposed to improve the efficiency of Transformer architecture. One major category is the sparse attention mechanism which reduces the number of attending pairs of tokens for the attention computation. 

~\\In this paper, we propose a novel method that introduces inductive bias from a graphical representation of a sequence. Our method results in both high accuracy and computational efficiency. We propose the masking of a sequence with its graphical representation using a sparse attention mechanism. Because our masked attention is applied using the edges from an AST that grow linearly as the sequence length increase, we decrease the computation complexity of the attention matrix from quadratic to linear. 

We propose the use of a graph diffusion mechanism to address the difficulty of modeling dependencies between isolated nodes in an input sequence. Combined with an approximation algorithm, our approach is equivalent to a multi-hop self-attention layer. We achieve higher accuracy while keeping a low computational cost.

We evaluated our model on two tasks, two code summarization datasets, and one variable misuse dataset. Two qualitative examples for code summarization are presented in Figure \ref{fig:cssamples}, more examples could be found in Appendix \ref{apdx:csexamples}. 
The results show that our model outperforms baselines in code summarization. On the variable misuse task, our model performs slightly below the state-of-the-art baselines in terms of accuracy but largely outperforms all other baselines. 
% We discuss potential reasons for the under-performance on this metric in Section \ref{res:var_misuse}. 
We analyze the computational efficiency of our model and show a significant advantage in both memory use and CPU inference time compared to all previous work. Our code, data, and models are publicly available in \url{https://github.com/chengjunyan1/Graph-Sparse-Transformer}.

\section{Related works}

One focus of previous deep learning methods for learning distributed representation of source code is fusing information from graph representations like AST, CFG, PDG (Program Dependency Graph). \citet{Code2Vec} proposed the use of random walks to learn an embedding for an AST, the AST embeddings are then concatenated with token embeddings and a context vector learned by an attention mechanism as input to a decoder. \citet{DeepCom} propose SBT to flatten an AST into a sequence which can then be used as input to a sequence model like LSTM. \citet{CGCNN} learned a CFG representation by DeepWalk \cite{deepwalk} with CNN and a sequence representation by LSTM, the representations then concatenated as the context vector. 

There are methods of exploiting graph information in sequence models. \citet{TreeLSTM} fused tree information into a sequence using a tree-structured LSTM. \citet{TextGCN} constructed a heterogeneous graph combining information from document and text which improves accuracy in the text classification task. \citet{TreeTransformer} incorporated the constituency tree into the computation of the value part of the self-attention layer which effectively conditions inductive bias on the attention computation.

%There are methods of exploiting graph information in sequence models.  \citet{TreeLSTM} fused tree information into a sequence using a tree-structured LSTM. \citet{TextGCN} constructed heterogeneous graphs of documents and text which improves accuracy in the text classification task. \citet{TreeTransformer} incorporated tree structures into the computation of the self-attention layer which effectively conditions inductive bias on the attention computation.

Transformer-based architectures recently became popular in multiple source code tasks. \citet{Ahmad} used a Transformer in the code summarization task, using the source code sequence as input to the model. They propose to use relative positional encoding and copy mechanism and improve the model accuracy when compared to a vanilla Transformer. \citet{GREAT} leveraged edge attribute information as the relation between tokens which are used as bias terms in the attention calculation.  \citet{CodeTransformer} incorporated AST structure as relative positional information and explored multilingual learning in the programming language discipline. 

Despite the success, a drawback of a Transformer architecture is the high computational cost of the self-attention mechanism. Several improvements in the Transformer architecture have been proposed. There are two main directions towards more efficient self-attention computation. The first direction is the sparsification of the attention computation which attends fewer pairs of tokens and reduces the computational cost \cite{SparseTransformer, RoutingTransformer, Bigbird}. Another direction focuses on the low-rank approximation of the attention matrix \cite{Reformer, Performer}. In this work, we concentrate on the first direction which is the sparsification of attention computation. \citet{EfficientTransformer} reviewed previous works around improving the efficiency of Transformers in detail.

\section{Method}

To solve the challenges of exploiting graph information and efficient computation, we combine a graph with the sparse attention computation in the masking mechanism. In this section, we firstly describe our graph sparse attention method in detail, then introduce the attention diffusion mechanism to address the problem of modeling dependencies between isolated nodes in a sequence. Finally, discuss the implementation problem in the sparse attention computation.

% In section \textit{Sparse Attention with Graph Information}
% % section \ref{sec:sparseattn}
% , we describe our graph sparse attention method in detail. In section \textit{Sparse Attention with Graph Diffusion}
% % section \ref{sec:graphdiffuse}
% we introduce the attention diffusion mechanism to address the problem of modeling dependencies between isolated nodes in a sequence. In section \textit{Implementation}
% % section \ref{sec:imple}
% we discuss the implementation problem in the sparse attention computation.

\subsection{Sparse Attention with Graph Information}
\label{sec:sparseattn}

~\\Sparsification of the attention computation is one major direction to improve the efficiency of Transformers. Sparse-attention achieved success in multiple disciplines, by improving the computational efficiency, and even improve the accuracy of the model \cite{SparseTransformer,Bigbird}. This method can be viewed as applying a mask on the attention matrix. The self-attention for one attention head can be written as 
\begin{equation}\label{eq:sparse_attention}
\begin{gathered}
h^{t+1}_i=W^O\sum_j m_{ij} A_{ij} W^Vh^t_j \\
A_{ij}=\sigma 
(W^Qh^t_i,W^Kh^t_j)
\end{gathered}
\end{equation}

where $h^t$ is the set of hidden states in layer $t$, $A$ is the attention weight matrix, $m$ is the mask matrix, $\sigma$ is the SoftMax function, $W^{Q} \in R^{d_{model} \times d_{k}}$, $W^{K} \in R^{d_{model} \times d_{k}}$, $W^{V} \in R^{d_{model} \times d_{v}}$, $W^{O} \in R^{H d_{k} \times d_{model}}$ are parameter matrices, and $H$ is the number of attention heads.  
When $m_{ij}=0$ the attention weight between nodes $i$ and $j$ will not be computed providing an efficiency advantage. The main challenge for applying sparse attention successfully is choosing a meaningful sparse mask. Correctly identifying which tokens we should attend to can create a lower sparsity mask while maintaining or improving accuracy. Different approaches have been proposed on deciding a good rule to generate a mask, like \citet{SparseTransformer}, \citet{RoutingTransformer}.

~\\The intuition behind sparse attention is that the learned attention matrix is usually sparse \cite{Reformer}. We can avoid wasting computational resources by masking low-interaction token pairs in the attention computation instead of letting the model learn to assign a low attention weight.
For programming languages, the graph representation of source code already provides clues as to which token pairs should be attended to during the attention computation. Representations like AST, CFG can depict a ``ground truth'' of the interactions between source code sequence tokens. This is because AST and CFG are exact and non-ambiguous graphs generated at a compiler-level. 
The distance between two nodes in a graph conveys information about the level of interaction between them during the program execution. The nodes that have direct edges between them represent strong interactions as opposed to nodes that are separated by multiple hops.

~\\\textbf{Graph Conditioned Sparse Mask}. Given a graph $\mathrm{G}=\{\gamma,V,E\}$ where $\gamma$ is the adjacency matrix, $V$ is the node-set, $E$ is the edge set. By replacing the fix-pattern mask matrix $m$ in eq.\ref{eq:sparse_attention} with the adjacency matrix $\gamma$, we embed the graph structure information into the self-attention computation mechanism.

From the recursive attention relationship in eq.\ref{eq:sparse_attention} we use the node attributes as the input hidden states $h^0$, where $h^0_i$ is the embedding for the attribute of node $V_i$. Following \citet{GREAT}, we use multiple edge attributes for each pair of nodes as a bias term added on the query vector $W^Qh_i$, defined as $W^E\sum_ke^k_{ij}$ where $k$ means the $k$-th edge between node $i$ and $j$, and $e^k_{ij}$ is the learned embedding for the attribute of edge $E^k_{ij}$. 

Following the methodology of \citet{GREAT}, for the variable misuse task, we evaluate on a multigraph of AST, CFG, and artificial graph structures introduced by the authors. %other edge types in conjunction. 

The resulting attention formula for a \textit{Graph Conditioned Sparse Mask} is

\begin{multline}
\label{eq:graph_conditioned_sparse_mask}
\begin{gathered}
h^{t+1}_i=W^O\sum_j \gamma_{ij} A_{ij} W^Vh^t_j \\
A_{ij}=\sigma (W^Qh^t_i+W^E\sum_ke^k_{ij},W^Kh^t_j) 
\end{gathered}
\end{multline}

%With the same sparsity, a graph-based attention mask contains more information than a random mask. 
With more information-dense graph representations, the model could achieve higher accuracy with lower sparsity than a mask with a weaker statistical bias between tokens. In our experiments, we evaluate such a claim by randomly generated masks. Using a mask that considers the interactions between all pairs of tokens can represent a complete graph. A sparse graph representation of the sequence can introduce isolation between nodes that don't interact which can help eliminate redundancy and noise. This improves the learned representation and the performance on downstream tasks, we do experiments for both a \textit{random mask} and a \textit{complete mask} and discuss our results in Section \ref{sec:anaattnmask}.

~\\\subsection{Sparse Attention with Graph Diffusion}
\label{sec:graphdiffuse}

\begin{figure}[h]
\subfigure[]{
\begin{minipage}[t]{0.47\columnwidth}
\label{F_ast_a}
\includegraphics[width=1\columnwidth]{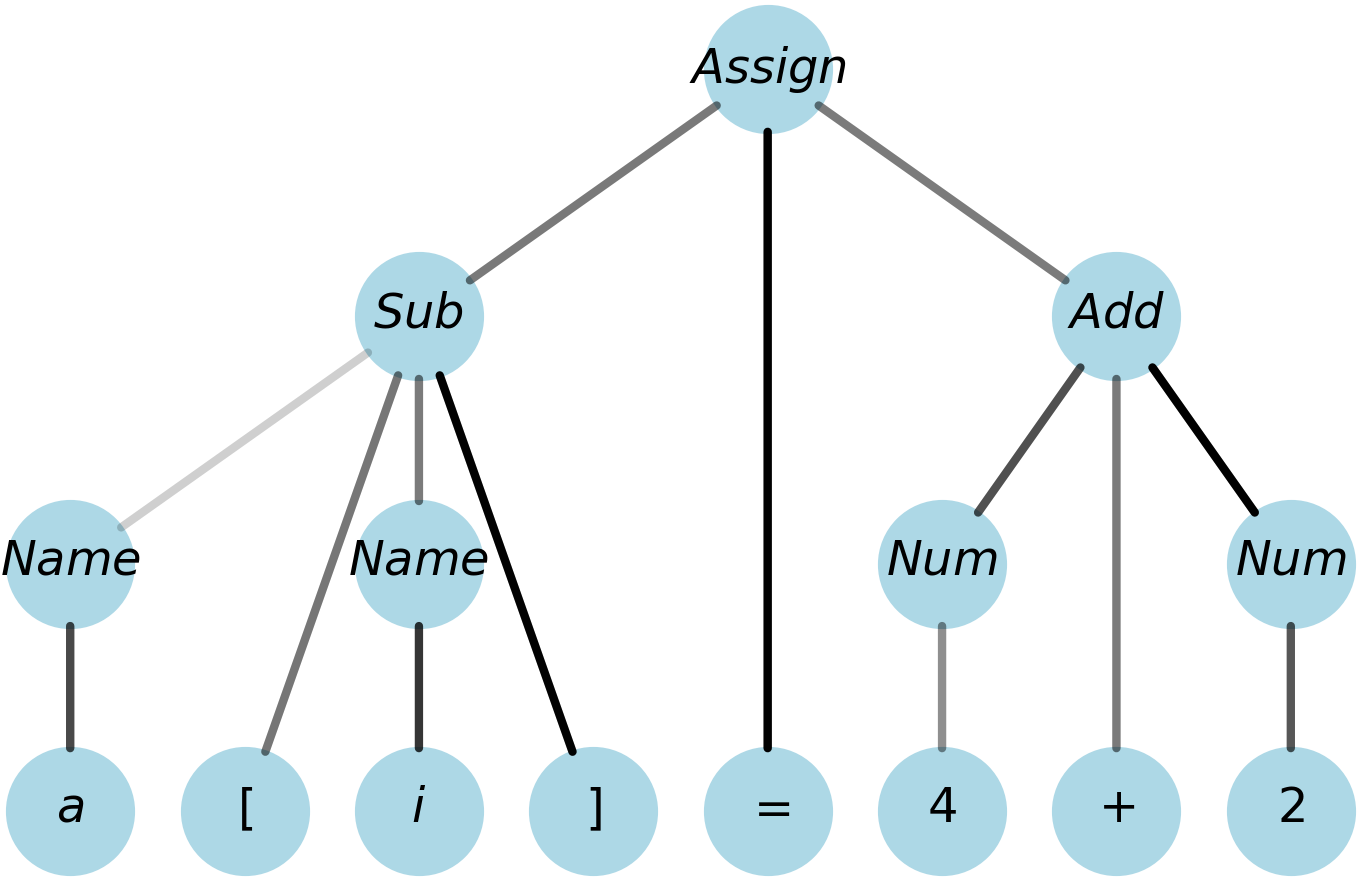}
\end{minipage}
}
\subfigure[]{
\begin{minipage}[t]{0.47\columnwidth}
\label{F_ast_b}
\includegraphics[width=1\columnwidth]{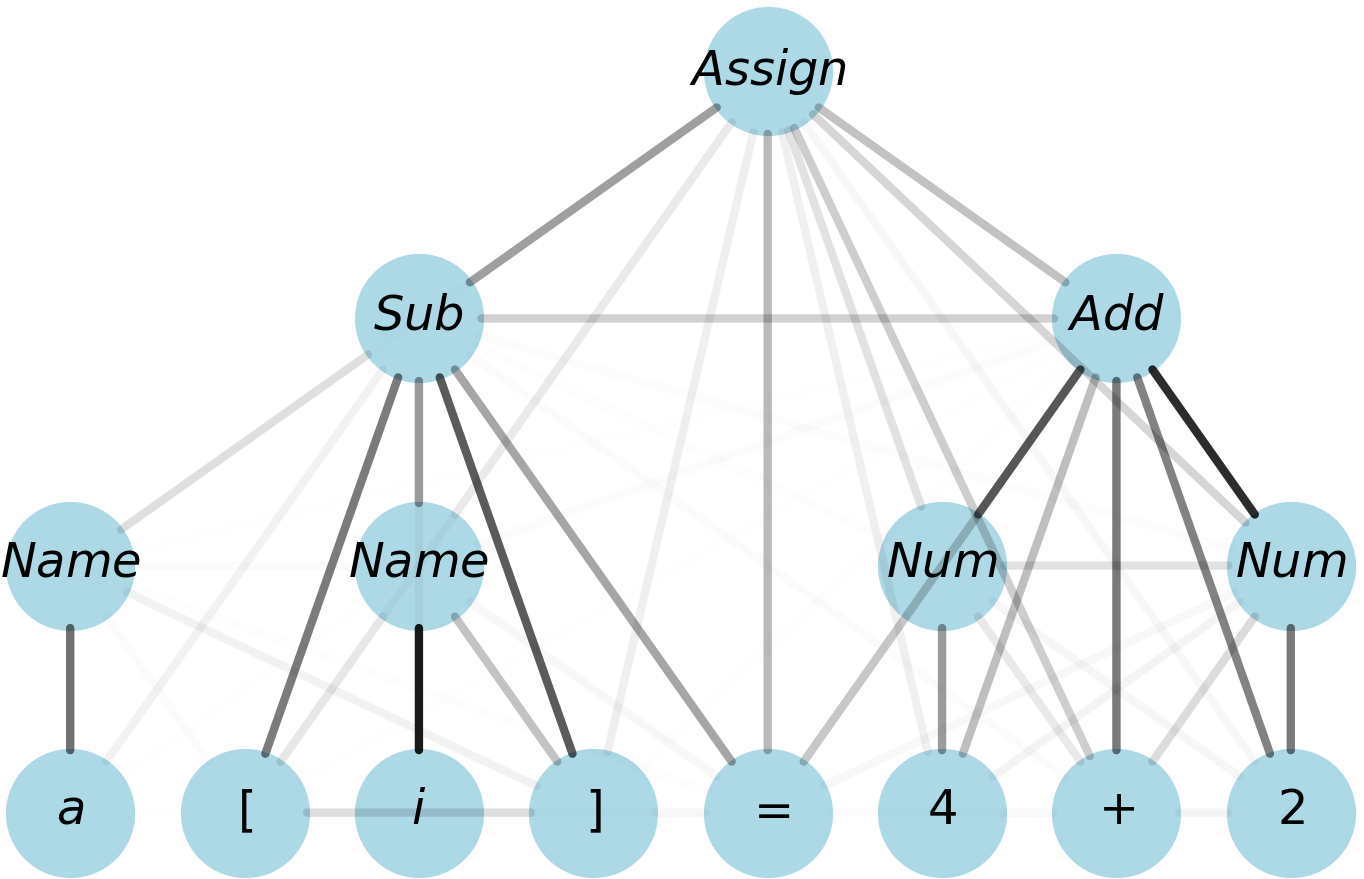}
\end{minipage}
}
\caption{(a) Attention weights visualized for an AST with a \textit{Graph Conditioned Sparse Mask}, the magnitude of transparency of an edge is proportional to the strength of attention weight between two nodes. (b) The same AST after \textit{Self-attention Diffusion Mechanism} where $K=2$.
}
\label{fig:astdoffise}
\end{figure}

~\\We propose \textbf{Self-attention Diffusion Mechanism} to effectively model the dependency of isolated tokens in a code sequence. To explain the need for a diffusion mechanism we need to introduce a graphical view of the attention mechanism. 

The Transformer could be regarded as executing a message-passing function on a complete graph. The Sparse-Attention mechanism is equivalent to a standard Transformer model for $m_{ij}=1$ in eq.\ref{eq:sparse_attention}. An attention layer performs a 1-hop message passing between each node and its neighboring nodes. In the case of a Transformer, each token can get global information from each other token due to the fully connected graph mask, thus a 1-hop message passing is enough to model the relationship between tokens. %The message passing between each attention layer of the model in both cases is 1-hop since for every layer each node is only considering its neighboring nodes. 

In the case of Sparse Attention, the attention graph is not fully connected, a 1-hop message passing mechanism has implications when two nodes have a strong relationship but are not directly connected. A code-based generated graph introduces isolation among nodes. There are naturally long-range dependencies in code that we can't effectively model in this way. Consider the AST in Figure \ref{F_ast_a} it would require 3-hops for message passing to occur from the leaf node `a’ to the root node `Assign’, which requires at least three attention layers. Thus for larger graphs, it would require a significantly larger model to learn effective relationships which is impractical.

~\\Graph diffusion \cite{GDC}, can help learn multi-hop information. We use the attention matrix as the transition matrix, the \textit{diffused sparse attention matrix} $\mathcal{D}$ is defined as:
\begin{center}
$\mathcal{D}=\sum_{i=0}^{\infty}\theta_i A^i$ where $\sum_{i=0}^{\infty}\theta_i=1, \theta_i>0$
\end{center}

The diffusion mechanism introduces no additional parameters to the model, however,
%can be counter to the idea of sparse attention. Adding new edges to the sparse masked attention counteracts the performance advantages that made us use this attention mechanism in the first place. 
calculating the exact diffusion matrix is intractable and can only be approximated. The diffusion mechanism can introduce new edges to the original graph and increase the computational cost. 

We adopt the Approximate Computation for Attention Diffusion by \citet{GDA} to solve this problem. We define the computation of \textit{Sparse Diffusion Mechanism} as: 
\begin{multline}\label{eq:sparse_diffusion}
\begin{gathered}
h^{t+1}=W^OZ^{(K)}, \lim_{K\to \infty } Z^{(K)} \approx \mathcal{D}W^Vh^t\\
Z^{(k+1)}=(1-\alpha)AZ^{(k)}+\alpha Z^{(0)} \\
\text{where}, Z^{(0)}=W^Vh^t
\end{gathered}
\end{multline}

We use $\alpha$ as a hyper-parameter that defines the strength of the diffusion mechanism. We approximate $ Z^{(K)}$  using a recurrence relationship with hyper-parameter $K$ that controls how many hops of messages a node could receive through diffusion. 

We found that a small $K=2$ is enough for getting a good approximation of $D$ which makes the computational cost of the diffusion mechanism small. We tested the computational cost of the diffusion mechanism for different $K$ in our experiments, the results are presented in Section \ref{resperf}.

~\\\subsection{Implementation}
\label{sec:imple}

Sparse computations are efficient for both inference and training time. The parallel computing techniques,  such as a GPU, currently used for training deep learning models cannot fully take advantage of sparse computations. It is an active research topic on how to improve the efficiency of sparse matrix operations on a GPU \cite{GPU1,GPU2}. There are existing optimization techniques like blockwise \cite{Bigbird}, that improve the computational efficiency of the sparse mask computation for a GPU. For our method, it is not applicable because such optimization techniques make assumptions on the structure of the attention mask. In contrast, a Graph Conditioned Sparse Mask that can display randomness created by the graph structures generated from AST. The insight we make agrees with our experimental results of similar training performance on a GPU between our approach and previous works.
 
~\\ In this work, we use an open-source library like DGL \cite{DGL} that provides optimized sparse computations on GPU and CPU. The computational advantages of GPU are limited since they are optimized for dense matrix computations. Any additional performance advantages are limited only to the optimization at the hardware or software level of sparse matrix computations.

\section{Experiments}
\begin{table*}[]
\centering
\caption{Results for Code Summarization task.}
\label{table:csresults}
\begin{tabular}{ccccccc}
\hline
            & \multicolumn{3}{c}{Java}                         & \multicolumn{3}{c}{Python}                       \\ \cline{2-7} 
            & BLEU           & METEOR         & ROUGE-L        & BLEU           & METEOR         & ROUGE-L        \\ \hline
DeepCom \cite{DeepCom}     & 39.75          & 23.06          & 52.67          & 20.78          & 9.98           & 37.35          \\
Dual Model \cite{DualModel}  & 42.39          & 25.77          & 53.61          & 21.80          & 11.14          & 39.45          \\
Transformer \cite{Transformer} & 43.41          & 25.91          & 52.71          & 31.08          & 18.57          & 44.31          \\
GREAT \cite{GREAT}       & 44.05          & 26.42          & 53.01          & 31.19          & 18.65          & 43.75          \\
C2NL \cite{Ahmad}        & 44.58          & 26.43          & 54.76          & 32.52          & 19.77          & 46.73 \\ \hline
Ours        & 45.21 & 26.55 & 55.00 & 33.31 & \textbf{19.82} & 46.56          \\
Ours (w/o D) & 44.36          & 26.06          & 54.49          & 33.02          & 19.65          & 46.43          \\
Ours (800 SL)        & \textbf{45.55} & \textbf{27.13} & \textbf{55.40} & \textbf{33.41} & 19.71 & \textbf{46.80}          \\ \hline
\end{tabular}
\end{table*}

\begin{table*}[!htb]
\centering
\caption{Results for Variable Misuse task.}
\label{table:vmresults}
\begin{tabular}{cccccc}
\hline
                 & Joint            & Bug-free         & Localization     & Repair           & Params \\ \hline
RNNs \cite{GREAT}            & 52.18\%          & 82.57\%          & 63.56\%          & 63.22\%          & 9.68M  \\
GGNNs \cite{GREAT}          & 65.38\%          & 90.28\%          & 79.64\%          & 75.76\%          & 41.19M \\
Sandwiches \cite{GREAT}       & 77.98\%          & 88.76\%          & 86.09\%          & 85.16\%          & 43.95M \\
Transformers \cite{Transformer}     & 66.05\%          & 91.70\%          & 73.39\%          & 76.79\%          & 26.22M \\
Transformers-10L \cite{Transformer} & 71.22\%          & 90.16\%          & 79.00\%          & 80.46\%          & 38.82M \\
GREAT \cite{GREAT}            & \textbf{78.21\%} & 88.98\%          & \textbf{86.14\%} & \textbf{85.85\%} & 26.22M \\ \hline
Ours             & 75.36\%          & 79.75\%          & 85.56\%          & 82.79\%          & 26.22M \\
Ours (w/o D)     & 70.53\%          & 73.24\%          & 84.50\%          & 78.93\%          & 26.22M \\
Ours-E           & 74.43\%          & \textbf{93.15\%} & \textbf{86.14\%} & 81.56\%          & 26.22M \\
Ours-E (w/o D)   & 68.87\%          & 92.93\%          & 84.39\%          & 77.37\%          & 26.22M \\ \hline
\end{tabular}
\end{table*}

~\\We evaluated our model in two tasks, Code summarization, and Variable misuse. Unless mentioned, all hyper-parameters used in our experiments are presented in appendix \ref{apdx:hyperparam}. 
% Our experiments for hyper-parameters presented in Appendix \ref{apdx:additional}.

For code summarization task, the goal is to generate a meaningful, human-readable summary of a given code snippet. On the variable misuse task, the goal is to predict whether a bug exists in a snippet and if it exists find the location of the two tokens in the code snippet that should be swapped. Specifically, there are two locations predicted, the first location corresponds to the misused variable (or a special value for ``no-bug'' classification), and the second location is variable that should be used to replace the misused one.

\subsection{Code summarization}
\label{sec:code_sum}

The experiments are conducted on a Java dataset \cite{TLCS} and a Python dataset \cite{PCSD}.
We used JavaParser\footnote{\url{https://javaparser.org/}} to extract the AST and javalang\footnote{\url{https://github.com/c2nes/javalang}} for parsing Java source code, python ast\footnote{\url{https://docs.python.org/3/library/ast.html}} to parse and get the AST for Python. 

For this task, we use a graph representation that connects each token in the source code only with its closest parent in AST.
AST nodes refer to non-leaf nodes in AST. A token may have multiple parents in an AST, but we only connect the token to the AST node with the deepest depth which is also its closest AST parent. All edges in the graph are bi-directional, and all nodes are self-looped. The above procedure results in the fusion of AST and a code sequence in one graph. The graph structure generated this way is closer to a tree, thus having sparse connections between nodes can improve the computation of the attention mask. We only provide the token representations as input for the decoder and discard the AST node representations. An illustration of the model can be found in Appendix \ref{apdx:modelstruct}.

\subsection{Variable misuse}
\label{exp:var_misuse}

% TODO: Iordanis stopped here
The experiments are conducted on the VarMisuse dataset from \citet{learn_code_graph}. We use positional encoding only for the bug-free classification task. 
The sequence of tokens is necessary for deciding whether an expression is valid or not. A positional encoding provides additional information that improves the model performance in this task. We found a shallow version of our model with positional encoding is adequate for learning to classify bug-free snippets. We used an ensemble model composed of a 5-layered version of our model without positional encoding and a 1-layered model with positional encoding in this task.

~\\\subsection{Details about Baselines}

We choose GREAT and a vanilla Transformer as our main baselines for both tasks, we also compared our model with other baselines in each task respectively. The code for preprocessing the raw source code files used by GREAT is not publicly available. In the code summarization task, we preprocessed both Java and Python dataset with the methodology outlined in section \ref{sec:code_sum} and used it to train a GREAT model. For the variable misuse task, we used the public preprocessed data provided by the authors of GREAT to train our model. 

%For a Transformer, we only use the source code sequence as input to the model, following the same methodology and model configuration as \citet{Ahmad}. 
In the code summarization task, we use the code publicly available by GREAT to train a model using identical data, model, and training configuration as our model. The results of other baselines are directly reported from \citet{Ahmad} for the code summarization task and \citet{GREAT} for the variable misuse task. Additional details on the baselines for each task are further described in Section \ref{resana}.

\subsection{Performance Analysis}

~\\We analyzed the performance of our model in memory use and CPU inference speed. Memory use corresponds to both train and test time memory use and can be a limiting factor on the training speed based on the effective batch size that can be used as well as the throughput during inference time. We compare our model with GREAT and Transformer, additional comparison with the remainder baselines can be found in Appendix \ref{apdx:perfdense}. 
We perform all our experiments in this section on an artificially generated dataset.

We generate random sequences in varying lengths, from 100 to 10000 tokens and by a step size of 100. For both the Graph Conditioned Masking and the Sparse Diffusion Masking we use randomly generated masks with triple the number of edges as the sequence length. 

The mask we use would be identical to a real dataset that uses an AST graph for the masking mechanism. The number of edges in an AST is the number of nodes minus one. Other artificial graph structures like the one used by \citet{GREAT} have the same linear relationship of edges to nodes in real source code datasets. Thus the edge number in both cases grows linearly with the sequence length. Additional analysis on the growth pattern of edges can be found in Appendix \ref{apdx:growthpattern}. 

%We also experimented on a real code snippet selected from the Java dataset. 
For consistency in the performance evaluation, all experiments are performed only with the encoder of the model. The representation extracted by an encoder can be used for any downstream task. Moreover, different models use different decoder architectures not allowing for a direct comparison of the attention mechanism.  %of since both our model and our baseline GREAT are focused on the encoder part moreover the decoder is varied for downstream tasks.

\subsection{Performance test on real code repositories}

\begin{table}[!htb]
\caption{Ablation studies on 10 real Java repositories. ``F'' represents directly using full sequence, remaining models are using truncated sequence with max length of 2000. ``D'' denotes our model with diffusion. Results shows the average CPU and GPU time (ms) and memory use (mb). ``Total'' denotes total file number used for test test.}
\label{table:perf_real}
\begin{tabular}{lccc}
\hline
Avg.       & CPU     & Mem     & GPU    \\ \hline
Ours (F)   & 1153.21 & 821.81  & 60.30  \\
Ours-D (F) & 1460.63 & 1003.81 & 74.55  \\ \hline
Ours       & 795.65  & 486.97  & 55.97  \\
Ours-D     & 872.95  & 617.93  & 67.45  \\
GREAT      & 1967.45 & 3148.06 & 102.74 \\
Trans.     & 910.51  & 1578.85 & 69.58  \\ \hline
Avg.       & Full    & Trunc.  &        \\
tokens     & 2133.32 & 1358.45 &        \\
nodes      & 2809.96 & 1760.45 &        \\ \hline
Total     & \multicolumn{3}{c}{28083}  \\ \hline
\end{tabular}
\end{table}

We performed performance test on 10 most popular Java repositories on GitHub including Flink, Eclipse Che, Elasticsearch, Hadoop, JDK1.8, libGDX, Presto, Spring Framework, IntelliJ Community, and Groovy. We cleared the non-java file and the non-class files like the files used as headers. Statistics and results could be found in Table \ref{table:perf_real}.

\section{Results and Analysis}
\label{resana}

\subsection{Code summarization results}
\label{sec:res_cod_sum}

Our method outperforms all baselines in Java and Python and in all metrics, See Table \ref{table:csresults}. The ``Ours'' refers to our full model with Graph Conditioned Sparse Masked attention and a Diffusion Mechanism. The ``Ours (w/o D)'' refers to our model without the Diffusion Mechanism. The ``Ours (800 SL)'' represents our full model trained with a larger maximum sequence length sequence of 800 instead of 150 and 400 for the Java and Python datasets. Our experiments demonstrate that longer sequence lengths greatly improve the accuracy of the model. All of the previous works only evaluate on a limited sequence length and are unable to capture long-range dependencies found in source code.  

%In this task, the diffusion mechanism also generally improved the performance of our model. 
As additional baselines, we report ``C2NL'' \cite{Ahmad} which is a vanilla Transformer equipped with the relative positional encoding and copy mechanism. ``DeepCom'' \cite{DeepCom} uses a structure-based traversal to flatten ASTs into sequences as input for a seq2seq model that consists of LSTMs. ``Dual Model'' \cite{DualModel} considers Code Summarization (CS) and Code Generation (CG) as a dual-task, it jointly trains a CS model and a CG model that consists of a BiLSTM encoder and an LSTM decoder.

\subsection{Variable misuse results}
\label{res:var_misuse}

The results are presented in Table \ref{table:vmresults}. "Ours-E" refers to the ensemble model as described in Section \ref{exp:var_misuse}. "Ours-E (w/o D)" is the ensemble model without a diffusion mechanism.  "Joint" refers to the joint bug localization and bug repair accuracy which is the main metric used in this task. "Localization" and "Repair" refer to bug localization and bug repair accuracy respectively. "Bug-free" refers to the accuracy of classifying the presence of a bug or not in a snippet. We refer readers to \citet{learn_code_graph} for more details about this task and the metrics. "RNNs'' refers to a 2-layer RNN, "GGNN'' is an 8-layer GGNN, "Sandwiches'' is a hybrid model which structure is "1R 4G 1R 4G 1R" where "1R'' represents 1 layer of RNN, "4G'' represents 4 layers of GGNN, sandwiches are the stack of these models.

%For this task, we directly used the public dataset from \citet{GREAT}, 

Since the data preprocessing for GREAT is not publicly available, we cannot make a dataset with a larger maximum sequence length as our methodology in the Code Summarization task. The results show that with the same number of parameters, the accuracy of our method is close to the state-of-the-art method as GREAT (about 4.83\% worse in the main metric). We still largely outperformed the previous state-of-the-art method Transformer by 12.69\% with the same number of parameters in the same task and metric. Given our observations from Section \ref{sec:res_cod_sum}, we anticipate that a model trained on a longer sequence length could increase the gap between our model performance and the baselines.

\subsection{Additional experiments for diffusion mechanism}
\label{apdx:additional}

\begin{table}[]
\caption{Additional experiments on Java dataset. ``L'' represents the number of layers. ``B.'', ``M.'', ``R.'' refers to BELU, METEOR and ROUGE-L respectively.}
\label{table:abl_java}
\begin{tabular}{ccc|ccc|c}
\hline
L       & K       & $\alpha$      & B.  & M. & R. & Param \\ \hline
3            & -       & -          & 38.86 & 20.54  & 50.11   & 22.1M   \\
3            & 2       & 0.25       & 39.40 & 20.86  & 50.51   & 22.1M   \\
3            & 3       & 0.25       & 39.81 & 21.25  & 50.71   & 22.1M   \\
3            & 4       & 0.25       & 40.17 & 21.41  & 51.11   & 22.1M   \\ \hline
6            & 2       & 0.10       & 44.87 & 26.37  & 54.68   & 44.1M   \\
6            & 2       & 0.15       & 44.99 & 26.37  & 54.92   & 44.1M   \\
6            & 2       & 0.25       & 45.21 & 26.55  & 55.00   & 44.1M   \\
6            & 2       & 0.35       & 44.88 & 26.31  & 54.64   & 44.1M   \\
6            & 3       & 0.25       &   45.33    &   55.12     &    26.64     & 44.1M   \\
6            & 4       & 0.25       & 45.38      &   55.22     &     26.70    & 44.1M   \\ \hline
10           & -       & -          & 45.17 & 26.76  & 55.02   & 73.6M   \\
10           & 2       & 0.25       & 46.22 & 27.97  & 55.81   & 73.6M   \\ \hline
\multicolumn{3}{c|}{2-hop attn.} & 43.90 & 25.24  & 53.49   & 50.4M   \\ \hline
\end{tabular}
\end{table}

We additional performed ablation study on $K$ and $\alpha$, and the gaining from diffusion on higher layer models, and the effect of $K$ on lower layer models. Results are listed on Table \ref{table:abl_java}. 
The results shows that in shallow model (3 layers) where the hops are low, increasing $K$ significantly improve performance. In a deeper model, increasing $K$ gives less improvement with a cost of efficiency, thus a $K=2$ gives good balance.
A too high or too low $\alpha$ are both not ideal. 
In deep model (10 layers), diffusion could still have gain. For model further deeper, there will be other problems like losing rank \cite{notallyouneed} which may influence the results and outside the scope of this paper. Such problems also reflected in our experiment on a 2-hop attention layer that stacking two attention layers in each model layer, which massage passing ability equivalent to a 6 layers model with $K=2$, result shows that 2-hop attention layer decreased the accuracy in contrary.

\subsection{Performance analysis}
\label{resperf}

% \usepackage{multirow}

\begin{figure}[h]
\subfigure[]{
\label{fig:perfmem}
\begin{minipage}[t]{0.45\textwidth}
\includegraphics[width=1.0\textwidth]{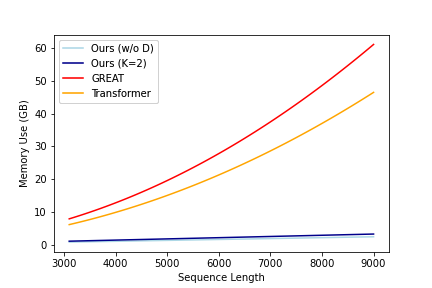}
\end{minipage}
}
\subfigure[]{
\label{fig:perfcpu}
\begin{minipage}[t]{0.45\textwidth}
\includegraphics[width=1.0\textwidth]{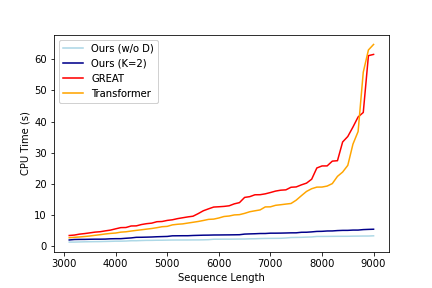}
\end{minipage}
}
\caption{Tests on (a) memory use (b) CPU inference time. % ``Dense'' represents when we use a dense matrix implementation to calculate the sparse attention matrix. ``w/o D'' represents without diffusion, 
}
\label{fig:perftext}
\end{figure}

Results on our artificial dataset are presented in Figure \ref{fig:perftext}. Figure \ref{fig:perfmem} shows our experiments on inference memory use. During training, memory use has the same growth trend as inference. We also implemented the calculation of sparse attention using a traditional dense matrix and observed identical growth rates as our baselines. Additional figures are in Appendix \ref{apdx:perfdense}. 
As the sequence length increased, dense attention computation methods reached a prohibitively high memory use. Such memory requirements firstly make training on larger sequence lengths intractable and secondly during inference doesn't allow for the deployment of these models to most user devices. For example, GREAT consumed more than 70GB of memory on the sequences with a length of 10000. With our sparse diffusion method we have a memory cost of less than 4GB for the same sequence length. Figure \ref{fig:perfcpu} shows the experimental results on CPU inference time and follows the same pattern as memory use. When the sequence length is 10000, GREAT used about 150s to encode the input, while our model with a diffusion mechanism only needed a fraction of the time. 

%Because our implementation makes use of DGL which also suffered from current challenge of implementing efficient sparse computation on GPU, the computational advantages are only limited in CPU. Hence we observed a similar quadratic growth pattern between dense attention matrix computation and our sparse approach. The limitation is only due to the hardware and the implementation details of DGL.

The performance advantage relies on sparsity of our attention matrix which is dependent on number of edges in the conditioning graph structure. For both AST and the GREAT dataset, edge number grows linearly with the sequence length. 
Additional performance results for real code snippets and Java repositories listed in Appendix \ref{apdx:realcode}.

~\\\subsection{Analysis on attention mask}
\label{sec:anaattnmask}

We perform an ablation study on the effect of attention mask on the code summarization task to verify our claim that the graph-based mask adds more information than a random mask or no mask at all. We used a ``Complete'' mask, which is a constant mask matrix of 1 as used by a Transformer, a ``Random'' mask with an edge density of 3\% which is identical to the density of the graph structure in our method. The results show that our method performs better than both a random mask with similar density and the complete graph in both summarization datasets. Moreover, the loss diverged in the Python dataset when a random mask was used. For the complete mask, it adds redundant connections which decrease the accuracy of the model due to the added noise. The full results can be found in Appendix \ref{apdx:attnmask}.

\section{Conclusion}

~\\In this work we propose a Graph Conditioned Sparse Mask Attention mechanism to introduce inductive bias and a Graph Diffusion Mechanism to address the isolation between nodes introduced by the sparse attention. We are able to model long-range dependencies of n-hop degree between graph nodes with no additional parameters or computational performance degradation. We outperform all baselines in terms of standardized metrics in the code summarization task and achieve near state-of-the-art accuracy in the variable misuse task. Our approach yields significant improvements in memory use and CPU inference time, reducing the growth of computational cost from quadratic to linear with respect to the sequence length.
We conclude that using our approach we can train and perform inference on longer sequences that can model long-range dependencies that are naturally present in source code. %This can lead to improved model performance in a downstream task. 
%The advantages of our method cannot be fully realized during training time because current state of the art in hardware is designed for dense matrix computation. 
Additional advances in sparse computation could help our method achieve better efficiency. Other future directions include better graph representations that have richer information than AST while matching the edge sparsity of AST.

% Use \bibliography{yourbibfile} instead or the References section will not appear in your paper
\bibliography{aaai22}

\begin{thebibliography}{31}
\providecommand{\natexlab}[1]{#1}

\bibitem[{Ahmad et~al.(2020)Ahmad, Chakraborty, Ray, and Chang}]{Ahmad}
Ahmad, W.~U.; Chakraborty, S.; Ray, B.; and Chang, K.-W. 2020.
\newblock A Transformer-based Approach for Source Code Summarization.
\newblock In \emph{Proceedings of the 58th Annual Meeting of the Association
  for Computational Linguistics (ACL)}.

\bibitem[{Allamanis, Brockschmidt, and Khademi(2017)}]{learn_code_graph}
Allamanis, M.; Brockschmidt, M.; and Khademi, M. 2017.
\newblock Learning to represent programs with graphs.
\newblock \emph{arXiv preprint arXiv:1711.00740}.

\bibitem[{Alon et~al.(2019)Alon, Zilberstein, Levy, and Yahav}]{Code2Vec}
Alon, U.; Zilberstein, M.; Levy, O.; and Yahav, E. 2019.
\newblock Code2vec: Learning Distributed Representations of Code.
\newblock \emph{Proc. ACM Program. Lang.}, 3(POPL).

\bibitem[{Arzt et~al.(2014)Arzt, Rasthofer, Fritz, Bodden, Bartel, Klein,
  Le~Traon, Octeau, and McDaniel}]{flowdroid}
Arzt, S.; Rasthofer, S.; Fritz, C.; Bodden, E.; Bartel, A.; Klein, J.;
  Le~Traon, Y.; Octeau, D.; and McDaniel, P. 2014.
\newblock Flowdroid: Precise context, flow, field, object-sensitive and
  lifecycle-aware taint analysis for android apps.
\newblock \emph{Acm Sigplan Notices}, 49(6): 259--269.

\bibitem[{Barone and Sennrich(2017)}]{PCSD}
Barone, A.; and Sennrich, R. 2017.
\newblock A parallel corpus of Python functions and documentation strings for
  automated code documentation and code generation.
\newblock In \emph{The 8th International Joint Conference on Natural Language
  Processing (IJCNLP 2017)}, volume~2, 314--319.

\bibitem[{Battaglia et~al.(2018)Battaglia, Hamrick, Bapst, Sanchez-Gonzalez,
  Zambaldi, Malinowski, Tacchetti, Raposo, Santoro, Faulkner et~al.}]{GN}
Battaglia, P.~W.; Hamrick, J.~B.; Bapst, V.; Sanchez-Gonzalez, A.; Zambaldi,
  V.; Malinowski, M.; Tacchetti, A.; Raposo, D.; Santoro, A.; Faulkner, R.;
  et~al. 2018.
\newblock Relational inductive biases, deep learning, and graph networks.
\newblock \emph{arXiv preprint arXiv:1806.01261}.

\bibitem[{Calcagno and Distefano(2011)}]{infer}
Calcagno, C.; and Distefano, D. 2011.
\newblock Infer: An automatic program verifier for memory safety of C programs.
\newblock In \emph{NASA Formal Methods Symposium}, 459--465. Springer.

\bibitem[{Child et~al.(2019)Child, Gray, Radford, and
  Sutskever}]{SparseTransformer}
Child, R.; Gray, S.; Radford, A.; and Sutskever, I. 2019.
\newblock Generating Long Sequences with Sparse Transformers.
\newblock arXiv:1904.10509.

\bibitem[{Choromanski et~al.(2020)Choromanski, Likhosherstov, Dohan, Song,
  Gane, Sarlos, Hawkins, Davis, Mohiuddin, Kaiser, Belanger, Colwell, and
  Weller}]{Performer}
Choromanski, K.; Likhosherstov, V.; Dohan, D.; Song, X.; Gane, A.; Sarlos, T.;
  Hawkins, P.; Davis, J.; Mohiuddin, A.; Kaiser, L.; Belanger, D.; Colwell, L.;
  and Weller, A. 2020.
\newblock Rethinking Attention with Performers.
\newblock arXiv:2009.14794.

\bibitem[{Dong, Cordonnier, and Loukas(2021)}]{notallyouneed}
Dong, Y.; Cordonnier, J.-B.; and Loukas, A. 2021.
\newblock Attention is Not All You Need: Pure Attention Loses Rank Doubly
  Exponentially with Depth.
\newblock arXiv:2103.03404.

\bibitem[{Dosovitskiy et~al.(2020)Dosovitskiy, Beyer, Kolesnikov, Weissenborn,
  Zhai, Unterthiner, Dehghani, Minderer, Heigold, Gelly
  et~al.}]{VisionTransformer}
Dosovitskiy, A.; Beyer, L.; Kolesnikov, A.; Weissenborn, D.; Zhai, X.;
  Unterthiner, T.; Dehghani, M.; Minderer, M.; Heigold, G.; Gelly, S.; et~al.
  2020.
\newblock An image is worth 16x16 words: Transformers for image recognition at
  scale.
\newblock \emph{arXiv preprint arXiv:2010.11929}.

\bibitem[{Gray, Radford, and Kingma(2017)}]{GPU1}
Gray, S.; Radford, A.; and Kingma, D.~P. 2017.
\newblock Gpu kernels for block-sparse weights.
\newblock \emph{arXiv preprint arXiv:1711.09224}, 3.

\bibitem[{Hellendoorn et~al.(2020)Hellendoorn, Sutton, Singh, Maniatis, and
  Bieber}]{GREAT}
Hellendoorn, V.~J.; Sutton, C.; Singh, R.; Maniatis, P.; and Bieber, D. 2020.
\newblock Global Relational Models of Source Code.
\newblock In \emph{International Conference on Learning Representations}.

\bibitem[{{Hu} et~al.(2018a){Hu}, {Li}, {Xia}, {Lo}, and {Jin}}]{DeepCom}
{Hu}, X.; {Li}, G.; {Xia}, X.; {Lo}, D.; and {Jin}, Z. 2018a.
\newblock Deep Code Comment Generation.
\newblock In \emph{2018 IEEE/ACM 26th International Conference on Program
  Comprehension (ICPC)}, 200--20010.

\bibitem[{Hu et~al.(2018b)Hu, Li, Xia, Lo, Lu, and Jin}]{TLCS}
Hu, X.; Li, G.; Xia, X.; Lo, D.; Lu, S.; and Jin, Z. 2018b.
\newblock Summarizing Source Code with Transferred API Knowledge.
\newblock In \emph{Proceedings of the Twenty-Seventh International Joint
  Conference on Artificial Intelligence, {IJCAI-18}}, 2269--2275. International
  Joint Conferences on Artificial Intelligence Organization.

\bibitem[{Huo, Li, and Zhou(2020)}]{CGCNN}
Huo, X.; Li, M.; and Zhou, Z.-H. 2020.
\newblock Control Flow Graph Embedding Based on Multi-Instance Decomposition
  for Bug Localization.
\newblock In \emph{AAAI}.

\bibitem[{Kitaev, Kaiser, and Levskaya(2020)}]{Reformer}
Kitaev, N.; Kaiser, L.; and Levskaya, A. 2020.
\newblock Reformer: The Efficient Transformer.
\newblock In \emph{International Conference on Learning Representations}.

\bibitem[{Klicpera, Wei\ss~enberger, and G\"{u}nnemann(2019)}]{GDC}
Klicpera, J.; Wei\ss~enberger, S.; and G\"{u}nnemann, S. 2019.
\newblock Diffusion Improves Graph Learning.
\newblock In Wallach, H.; Larochelle, H.; Beygelzimer, A.; d\textquotesingle
  Alch\'{e}-Buc, F.; Fox, E.; and Garnett, R., eds., \emph{Advances in Neural
  Information Processing Systems}, volume~32, 13354--13366. Curran Associates,
  Inc.

\bibitem[{Nguyen et~al.(2020)Nguyen, Joty, Hoi, and Socher}]{TreeTransformer}
Nguyen, X.-P.; Joty, S.; Hoi, S.; and Socher, R. 2020.
\newblock Tree-Structured Attention with Hierarchical Accumulation.
\newblock In \emph{International Conference on Learning Representations}.

\bibitem[{Perozzi, Al-Rfou, and Skiena(2014)}]{deepwalk}
Perozzi, B.; Al-Rfou, R.; and Skiena, S. 2014.
\newblock DeepWalk: Online Learning of Social Representations.
\newblock In \emph{Proceedings of the 20th ACM SIGKDD International Conference
  on Knowledge Discovery and Data Mining}, KDD '14, 701--710. New York, NY,
  USA: ACM.
\newblock ISBN 978-1-4503-2956-9.

\bibitem[{Roy et~al.(2020)Roy, Saffar, Vaswani, and
  Grangier}]{RoutingTransformer}
Roy, A.; Saffar, M.; Vaswani, A.; and Grangier, D. 2020.
\newblock Efficient Content-Based Sparse Attention with Routing Transformers.
\newblock arXiv:2003.05997.

\bibitem[{Tai, Socher, and Manning(2015)}]{TreeLSTM}
Tai, K.~S.; Socher, R.; and Manning, C.~D. 2015.
\newblock Improved Semantic Representations From Tree-Structured Long
  Short-Term Memory Networks.
\newblock In \emph{Proceedings of the 53rd Annual Meeting of the Association
  for Computational Linguistics and the 7th International Joint Conference on
  Natural Language Processing (Volume 1: Long Papers)}, 1556--1566. Beijing,
  China: Association for Computational Linguistics.

\bibitem[{Tay et~al.(2020)Tay, Dehghani, Bahri, and
  Metzler}]{EfficientTransformer}
Tay, Y.; Dehghani, M.; Bahri, D.; and Metzler, D. 2020.
\newblock Efficient Transformers: A Survey.
\newblock arXiv:2009.06732.

\bibitem[{Vaswani et~al.(2017)Vaswani, Shazeer, Parmar, Uszkoreit, Jones,
  Gomez, Kaiser, and Polosukhin}]{Transformer}
Vaswani, A.; Shazeer, N.; Parmar, N.; Uszkoreit, J.; Jones, L.; Gomez, A.~N.;
  Kaiser, u.; and Polosukhin, I. 2017.
\newblock Attention is All You Need.
\newblock In \emph{Proceedings of the 31st International Conference on Neural
  Information Processing Systems}, NIPS'17, 6000–6010. Red Hook, NY, USA:
  Curran Associates Inc.
\newblock ISBN 9781510860964.

\bibitem[{Wang et~al.(2020)Wang, Ying, Huang, and Leskovec}]{GDA}
Wang, G.; Ying, R.; Huang, J.; and Leskovec, J. 2020.
\newblock Direct Multi-hop Attention based Graph Neural Network.
\newblock arXiv:2009.14332.

\bibitem[{Wang et~al.(2019)Wang, Zheng, Ye, Gan, Li, Song, Zhou, Ma, Yu, Gai,
  Xiao, He, Karypis, Li, and Zhang}]{DGL}
Wang, M.; Zheng, D.; Ye, Z.; Gan, Q.; Li, M.; Song, X.; Zhou, J.; Ma, C.; Yu,
  L.; Gai, Y.; Xiao, T.; He, T.; Karypis, G.; Li, J.; and Zhang, Z. 2019.
\newblock Deep Graph Library: A Graph-Centric, Highly-Performant Package for
  Graph Neural Networks.
\newblock \emph{arXiv preprint arXiv:1909.01315}.

\bibitem[{Wei et~al.(2019)Wei, Li, Xia, Fu, and Jin}]{DualModel}
Wei, B.; Li, G.; Xia, X.; Fu, Z.; and Jin, Z. 2019.
\newblock Code Generation as Dual Task of Code Summarization.
\newblock \emph{NeurIPS}, 6559--6569.

\bibitem[{Yao, Mao, and Luo(2019)}]{TextGCN}
Yao, L.; Mao, C.; and Luo, Y. 2019.
\newblock Graph convolutional networks for text classification.
\newblock In \emph{Proceedings of the AAAI Conference on Artificial
  Intelligence}, volume~33, 7370--7377.

\bibitem[{Yao et~al.(2019)Yao, Cao, Xiao, Zhang, and Nie}]{GPU2}
Yao, Z.; Cao, S.; Xiao, W.; Zhang, C.; and Nie, L. 2019.
\newblock Balanced Sparsity for Efficient DNN Inference on GPU.
\newblock \emph{Proceedings of the AAAI Conference on Artificial Intelligence},
  33: 5676--5683.

\bibitem[{Zaheer et~al.(2020)Zaheer, Guruganesh, Dubey, Ainslie, Alberti,
  Ontanon, Pham, Ravula, Wang, Yang et~al.}]{Bigbird}
Zaheer, M.; Guruganesh, G.; Dubey, K.~A.; Ainslie, J.; Alberti, C.; Ontanon,
  S.; Pham, P.; Ravula, A.; Wang, Q.; Yang, L.; et~al. 2020.
\newblock Big bird: Transformers for longer sequences.
\newblock \emph{Advances in Neural Information Processing Systems}, 33.

\bibitem[{Z{\"u}gner et~al.(2021)Z{\"u}gner, Kirschstein, Catasta, Leskovec,
  and G{\"u}nnemann}]{CodeTransformer}
Z{\"u}gner, D.; Kirschstein, T.; Catasta, M.; Leskovec, J.; and G{\"u}nnemann,
  S. 2021.
\newblock Language-Agnostic Representation Learning of Source Code from
  Structure and Context.
\newblock In \emph{International Conference on Learning Representations}.

\end{thebibliography}

\clearpage
\appendix

\section{Summary of Hyper-parameters}
\label{apdx:hyperparam}

\begin{table}[H]
\begin{center}
\begin{tabular}{lll}
\multicolumn{1}{c}{\bf Base Model Config}  &\multicolumn{1}{c}{\bf Value} 
\\ \hline \\
Num of layers           &6\\
Attention heads         &8\\
$d_{k}$                 &64\\
$d_{v}$                 &64\\
$d_{model}$             &512\\
$d_{ff}$                &2048\\
$K$                     &2\\
$\alpha$                &0.25\\
\end{tabular}
\end{center}
\end{table}

\begin{table}[H]
\begin{center}
\begin{tabular}{lll}
\multicolumn{1}{c}{\bf Train (Code Sum.)}  &\multicolumn{1}{c}{\bf Value} 
\\ \hline \\
Dropout rate            &0.2\\
Optimizer               &Adam\\
Learning rate           &0.0001\\
Decay rate              &0.99\\
Max epoch num           &200\\
Early stop epochs       &20\\
Training Batch set      &30\\
Testing Beam size       &4\\
Max src. vocab size     &50000\\
Max tgt. vocab size     &30000\\
Max code len. (J)       &150\\
Max code len. (P)       &400\\
Max sum. len. (J)       &50\\
Max sum. len. (P)       &30\\
\end{tabular}
\end{center}
\end{table}

\begin{table}[H]
\begin{center}
\begin{tabular}{lll}
\multicolumn{1}{c}{\bf Train (Var. Misuse)}  &\multicolumn{1}{c}{\bf Value} 
\\ \hline \\
Dropout rate            &0.1\\
Optimizer               &Adam\\
Batch size              &7500\\
Learning rate           &0.0001\\
Max code len.:          &512\\
\end{tabular}
\caption{
The hyperparameters are consistent with \citet{Ahmad} for the code summarization task and \citet{GREAT} for the variable misuse task. 
}
\end{center}
\end{table}

\section{Investigation on GitHub projects}
\label{apdx:gitproj}

\begin{table}[H]
\begin{center}
\begin{tabular}{lcc}
\multicolumn{1}{c}{\textbf{Project}} & \textbf{Total lines} & \textbf{Avg lines} \\ \hline
Flutter                              & 911261               & 269.6              \\
Go lang.                             & 1696067              & 220.7              \\
Kubernetes                           & 4236562              & 265.8              \\
d3                                   & 2672                 & 296.9              \\
Deno                                 & 309239               & 228.2              \\
React                                & 327844               & 181.2              \\
Tensorflow                           & 2894245              & 208.4              \\
VScode                               & 1068910              & 245.3              \\
Vue                                  & 142670               & 266.7              \\
Linux                                & 21163798             & 349.4             
\end{tabular}
\caption{The projects are selected from the most starred projects in GitHub. Blank lines and comments are omitted when counting the number of lines, only code files are counted. All project have about 200 lines of code per file, and one line of code could contain multiple tokens, and any single file may contains more than 200 lines of code. Many software projects could easily exceed the sequence length limit of a Transformers architecture}
\end{center}
\end{table}

\section{Additional Performance test with dense implementation}
\label{apdx:perfdense}

\begin{figure}[H]
\centering
\subfigure[]{
\begin{minipage}[t]{0.8\columnwidth}
\label{F_node_edge_a}
\includegraphics[width=1\columnwidth]{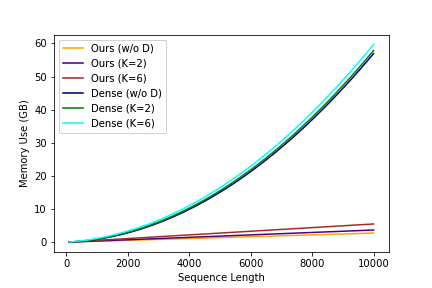}
\end{minipage}
}
\subfigure[]{
\begin{minipage}[t]{0.8\columnwidth}
\label{F_node_edge_b}
\includegraphics[width=1\columnwidth]{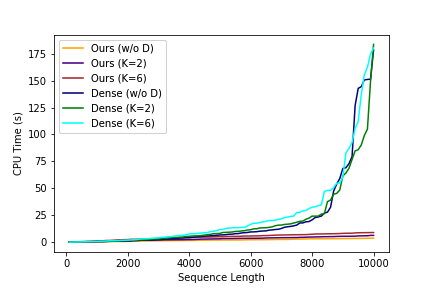}
\end{minipage}
}
\caption{We perform an additional experiment and compare the sparse implementation and dense implementation of our method in the growth of (a) memory use and (b) CPU time with respect to the sequence length.}
\label{F_perf2}
\end{figure}

\section{Model structure for Code summarization}
\label{apdx:modelstruct}

\begin{figure}[h]
\centering
\includegraphics[width=1\columnwidth]{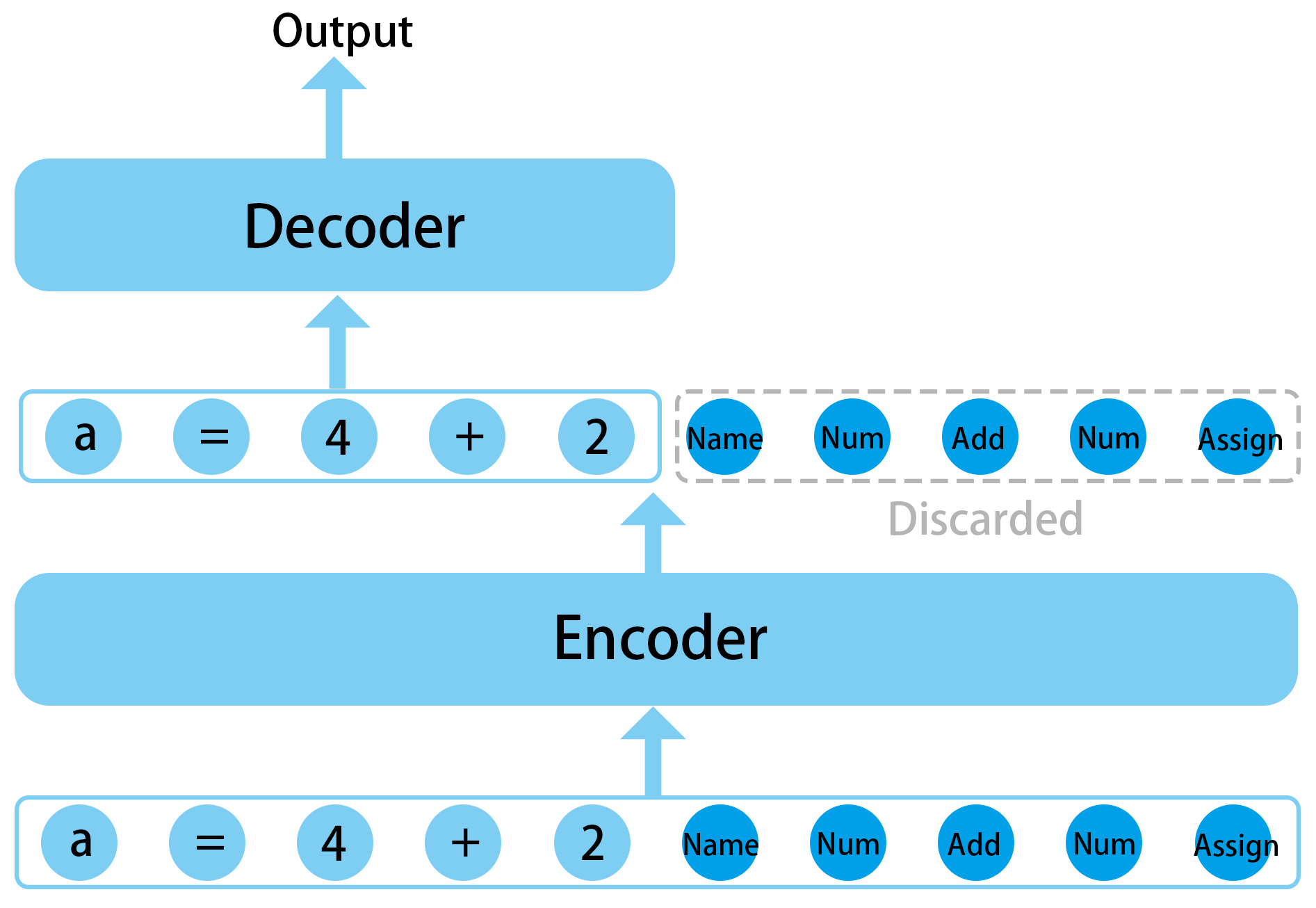}
\caption{The seq2seq architecture we used for the code summarization task. The AST node representations are discarded in the decoding stage.}
\end{figure}

%The seq2seq architecture we used for code summarization task. The AST node representations are discarded in the decoding stage.

% \clearpage

\section{Growth pattern of node and edge number}
\label{apdx:growthpattern}

\begin{figure}[H]
\centering
\subfigure[]{
\begin{minipage}[t]{0.8\columnwidth}
\label{F_node_edge_a}
\includegraphics[width=1\columnwidth]{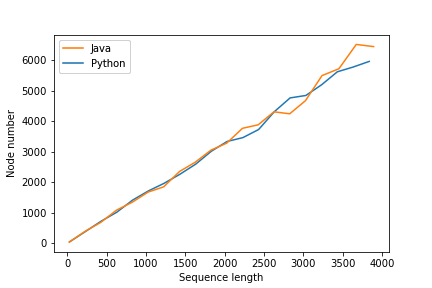}
\end{minipage}
}
\subfigure[]{
\begin{minipage}[t]{0.8\columnwidth}
\label{F_node_edge_b}
\includegraphics[width=1\columnwidth]{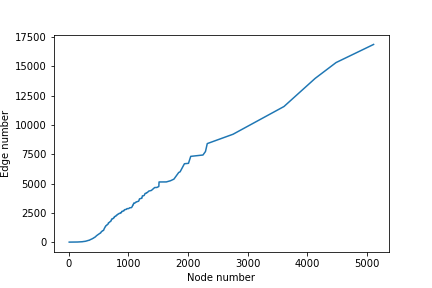}
\end{minipage}
}
\caption{We analyzed the growth pattern of sequence length with the respect to the edge number (a) The number of edges grows linearly with the sequence length in an AST. (b) The edge number also grows linearly with respect to the node number in GREAT's graph structure.}
\label{F_great_s}
\end{figure}

\section{Experiments on Attention mask}
\label{apdx:attnmask}

\begin{table}[H]
\centering
\begin{tabular}{clccc}
\hline
                              &            & BLEU     & METEOR   & ROUGE-L  \\ \hline
\multirow{2}{*}{\begin{tabular}[c]{@{}c@{}}Ours\\ (w/o D)\end{tabular}}  & \textit{J} & 44.36    & 26.06    & 54.49    \\
                              & \textit{P} & 33.02    & 19.65    & 46.43    \\ \hline
\multirow{2}{*}{FC}     & \textit{J} & 44.23    & 26.66    & 53.46    \\
                              & \textit{P} & 31.06    & 18.34    & 43.37    \\ \hline
\multirow{2}{*}{RD} & \textit{J} & 43.24    & 25.14    & 52.33    \\
                              & \textit{P} & Diverged & Diverged & Diverged \\ \hline
\end{tabular}
\caption{``RD'' represents random mask with 3\% sparsity, ``FC'' is a complete graph mask, ``J'' and ``P'' refers to Java and Python dataset respectively. Results show that our graph based method outperform both random and complete graphs in the two dataset and in all metrics.}
\end{table}

\section{Performance test on largest snippet in Java dataset}
\label{apdx:realcode}
\begin{table}[H]
\begin{center}
\begin{tabular}{ccccc}
\hline
\multicolumn{1}{c|}{Model}                                                    & \multicolumn{1}{c|}{\begin{tabular}[c]{@{}c@{}}Mem\\ (GB)\end{tabular}} & \multicolumn{1}{c|}{\begin{tabular}[c]{@{}c@{}}CPU\\ (ms)\end{tabular}} & \multicolumn{1}{c|}{\begin{tabular}[c]{@{}c@{}}GPU\\ (ms)\end{tabular}} & \begin{tabular}[c]{@{}c@{}}Seq\\ Len.\end{tabular}                     \\ \hline
\multicolumn{5}{c}{Original}                                                                                                                                                                                                                                                                                                                                                         \\ \hline
\multicolumn{1}{c|}{\begin{tabular}[c]{@{}c@{}}Sparse (w/o D)\end{tabular}} & \multicolumn{1}{c|}{2417}                                               & \multicolumn{1}{c|}{2417}                                               & \multicolumn{1}{c|}{103}                                                & \multirow{4}{*}{\begin{tabular}[c]{@{}c@{}}5956\\ (8802)\end{tabular}} \\
\multicolumn{1}{c|}{\begin{tabular}[c]{@{}c@{}}Sparse (K=2)\end{tabular}}   & \multicolumn{1}{c|}{3253}                                               & \multicolumn{1}{c|}{3761}                                               & \multicolumn{1}{c|}{121}                                                &                                                                        \\
\multicolumn{1}{c|}{\begin{tabular}[c]{@{}c@{}}Sparse (K=6)\end{tabular}}   & \multicolumn{1}{c|}{4905}                                               & \multicolumn{1}{c|}{6695}                                               & \multicolumn{1}{c|}{148}                                                &                                                                        \\ \cline{1-4}
\multicolumn{1}{c|}{Other}                                                    & \multicolumn{1}{c|}{OOM}                                                & \multicolumn{1}{c|}{OOM}                                                & \multicolumn{1}{c|}{OOM}                                                &                                                                        \\ \hline
\multicolumn{5}{c}{Truncated}                                                                                                                                                                                                                                                                                                                                                        \\ \hline
\multicolumn{1}{c|}{\begin{tabular}[c]{@{}c@{}}Sparse (w/o D)\end{tabular}} & \multicolumn{1}{c|}{819}                                                & \multicolumn{1}{c|}{847}                                                & \multicolumn{1}{c|}{73}                                                 & \multirow{8}{*}{\begin{tabular}[c]{@{}c@{}}2000\\ (3014)\end{tabular}} \\
\multicolumn{1}{c|}{\begin{tabular}[c]{@{}c@{}}Sparse (K=2)\end{tabular}}   & \multicolumn{1}{c|}{1106}                                               & \multicolumn{1}{c|}{1316}                                               & \multicolumn{1}{c|}{96}                                                 &                                                                        \\
\multicolumn{1}{c|}{\begin{tabular}[c]{@{}c@{}}Sparse (K=6)\end{tabular}}   & \multicolumn{1}{c|}{1674}                                               & \multicolumn{1}{c|}{2314}                                               & \multicolumn{1}{c|}{134}                                                &                                                                        \\ \cline{1-4}
\multicolumn{1}{c|}{\begin{tabular}[c]{@{}c@{}}Dense (w/o D)\end{tabular}}  & \multicolumn{1}{c|}{6015}                                               & \multicolumn{1}{c|}{6177}                                               & \multicolumn{1}{c|}{120}                                                &                                                                        \\
\multicolumn{1}{c|}{\begin{tabular}[c]{@{}c@{}}Dense (K=2)\end{tabular}}    & \multicolumn{1}{c|}{6298}                                               & \multicolumn{1}{c|}{6890}                                               & \multicolumn{1}{c|}{194}                                                &                                                                        \\
\multicolumn{1}{c|}{\begin{tabular}[c]{@{}c@{}}Dense (K=6)\end{tabular}}    & \multicolumn{1}{c|}{6873}                                               & \multicolumn{1}{c|}{7952}                                               & \multicolumn{1}{c|}{256}                                                &                                                                        \\ \cline{1-4}
\multicolumn{1}{c|}{GREAT}                                                    & \multicolumn{1}{c|}{7732}                                               & \multicolumn{1}{c|}{4532}                                               & \multicolumn{1}{c|}{535}                                                &                                                                        \\
\multicolumn{1}{c|}{Transformer}                                                   & \multicolumn{1}{c|}{6028}                                               & \multicolumn{1}{c|}{1768}                                               & \multicolumn{1}{c|}{119}                                                &                                                                        \\ \hline
\end{tabular}
\caption{Performance test on a real code snippet. ``Sparse'' and ``Dense'' represents sparse and dense implementation of our method respectively, ``Trans.'' is vanilla Transformer. Unlike other methods in the table, we do not use graph information from the sequence for the vanilla Transformer. In the Sequence Length column, the number in the parenthesis represents the total number of nodes (token number plus AST node number) for the sample. 
}
\end{center}
\end{table}

\section{Additional code summarization examples}
\label{apdx:csexamples}

\subsection{Examples from test set}

~\\
\begin{lstlisting} [language = Java,
        keywordstyle=\color{blue!70},
        commentstyle=\color{red!50!green!50!blue!50},
        frame=topline|bottomline|leftline|rightline,
        rulesepcolor=\color{red!20!green!20!blue!20},
        basicstyle=\small ]
public final boolean isExceptionHandlerEquivalent(BasicBlock other){
    if (exceptionHandlers != other.exceptionHandlers) {
        Enumeration<BasicBlock> e1=getExceptionHandlers();
        Enumeration<BasicBlock> e2=other.getExceptionHandlers();
        while (e1.hasMoreElements()) {
            if (!e2.hasMoreElements()) return false;
            if (e1.nextElement() != e2.nextElement()) return false;
        }
        if (e2.hasMoreElements()) return false;
    }
    return true;
}
\end{lstlisting}

\textbf{Reference:} compare the in scope exception handlers of two blocks.

\textbf{Ours:} return true if the block is equivalent to the exception handlers, or false otherwise.

\textbf{GREAT:} convert this block to another objects .

~\\
\begin{lstlisting} [language = Python,
        keywordstyle=\color{blue!70},
        commentstyle=\color{red!50!green!50!blue!50},
        frame=topline|bottomline|leftline|rightline,
        rulesepcolor=\color{red!20!green!20!blue!20},
        basicstyle=\small]
def issues_closed_since(period=timedelta(days=365), project=`ipython/ipython', pulls=False): 
    which = (`pulls' if pulls else `issues') 
    if isinstance(period, timedelta): 
        since = round_hour((datetime.utcnow() - period)) 
    else: 
        since = period 
    url = (`https://api.github.com/repos/%s/%s?
    state=closed&sort=updated&since=%s&per_page=%i' % (project, which, since.strftime(ISO8601), PER_PAGE)) 
    allclosed = get_paged_request(url, headers=make_auth_header()) 
    filtered = [i for i in allclosed if (_parse_datetime(i[`closed_at']) > since)] 
    if pulls: 
        filtered = [i for i in filtered if (_parse_datetime(i[`merged_at']) > since)] 
        filtered = [i for i in filtered if (i[`base'][`ref'] == `master')] 
    else: 
        filtered = [i for i in filtered if (not is_pull_request(i))] 
    return filtered
\end{lstlisting}

\textbf{Reference:} get all issues closed since a particular point in time .

\textbf{Ours:} return a list of closed issues .

\textbf{GREAT:} pull requests iso unreferenced for single-line issues .

\subsection{Examples from real snippets}

Here we show the summarizations our model generated for two real snippets chosen from large Java projects.

\begin{lstlisting} [language = Java,
        keywordstyle=\color{blue!70},
        commentstyle=\color{red!50!green!50!blue!50},
        frame=topline|bottomline|leftline|rightline,
        rulesepcolor=\color{red!20!green!20!blue!20},
        basicstyle=\small ]
  protected void write(String toWrite) throws IOException  {
    if(!okToWrite)
      throw new IOException("file not open for writing.");

    if(printToScreen)
      System.out.print(toWrite);

    try {
      fw.write(toWrite);
    } catch (IOException e) {
      okToWrite = false;
      throw e;
    }
  }
\end{lstlisting}

\textbf{Original file}: org.apache.hadoop.hdfs.tools.
offlineImageViewer.TextWriterImageVisitor

\textbf{Summary}: Write the image data to a file.

\begin{lstlisting} [language = Java,
        keywordstyle=\color{blue!70},
        commentstyle=\color{red!50!green!50!blue!50},
        frame=topline|bottomline|leftline|rightline,
        rulesepcolor=\color{red!20!green!20!blue!20},
        basicstyle=\small ]
public class StringTupleDeserializerMap implements MapFunction<byte[], Tuple1<String>> {
	@Override
	public Tuple1<String> map(byte[] value) throws Exception {
		return new Tuple1<>(new String(value, 5, value.length - 5, ConfigConstants.DEFAULT_CHARSET));
	}
}
\end{lstlisting}

\textbf{Original file}: org.apache.flink.python.api.
functions.util.StringTupleDeserializerMap

\textbf{Summary}: Parses the given string representation of a tuple.

\end{document}